\documentclass[conference, a4paper]{IEEEtran}
\IEEEoverridecommandlockouts
\usepackage{cite}
\usepackage{amsmath,amssymb,amsfonts}
\usepackage{algorithmic}
\usepackage{graphicx}
\usepackage{textcomp}

\usepackage{multirow}
\usepackage{float}
\usepackage{subfigure}
\usepackage{placeins}
\usepackage{array}
\usepackage{pifont}

\usepackage{xcolor}
\def\BibTeX{{\rm B\kern-.05em{\sc i\kern-.025em b}\kern-.08em
    T\kern-.1667em\lower.7ex\hbox{E}\kern-.125emX}}
\begin{document}

\title{Dynamic Attention and Bi-directional Fusion for Safety Helmet Wearing Detection\\

}

\author{
\IEEEauthorblockN{1\textsuperscript{st} Junwei Feng}
\IEEEauthorblockA{\textit{School of Astronautics} \\
\textit{Northwestern Polytechnical University}\\
Xi'an, China \\
fjw@mail.nwpu.edu.cn}
\and
\IEEEauthorblockN{2\textsuperscript{nd} Xueyan Fan}
\IEEEauthorblockA{\textit{School of Software} \\
\textit{Northwestern Polytechnical University}\\
Xi'an, China \\
fanxueyan@mail.nwpu.edu.cn}
\and
\IEEEauthorblockN{3\textsuperscript{rd} Yuyang Chen}
\IEEEauthorblockA{\textit{School of Software} \\
\textit{Northwestern Polytechnical University}\\
Xi'an, China \\
cyys@mail.nwpu.edu.cn}
\and
\IEEEauthorblockN{4\textsuperscript{th} Yi Li}
\IEEEauthorblockA{\begin{minipage}{\linewidth}\centering
\textit{Unmanned System Research Institute} \\
\textit{Northwestern Polytechnical University}\\
Xi'an, China \\
lyii@mail.nwpu.edu.cn
\end{minipage}}
}


\maketitle

\begin{abstract}
Ensuring construction site safety requires accurate and real-time detection of workers' safety helmet use, despite challenges posed by cluttered environments, densely populated work areas, and hard-to-detect small or overlapping objects caused by building obstructions. This paper proposes a novel algorithm for safety helmet wearing detection, incorporating a dynamic attention within the detection head to enhance multi-scale perception. The mechanism combines feature-level attention for scale adaptation, spatial attention for spatial localization, and channel attention for task-specific insights, improving small object detection without additional computational overhead. Furthermore, a two-way fusion strategy enables bidirectional information flow, refining feature fusion through adaptive multi-scale weighting, and enhancing recognition of occluded targets. Experimental results demonstrate a 1.7\% improvement in mAP@[.5:.95] compared to the best baseline while reducing GFLOPs by 11.9\% on larger sizes. The proposed method surpasses existing models, providing an efficient and practical solution for real-world construction safety monitoring.

\end{abstract}


\section{Introduction}
With the rapid growth of industry, ensuring worker safety has become a critical priority. Safety helmets\cite{1,2,3,8,48} have been shown to significantly reduce injury risks; however, many workers fail to wear helmets consistently on-site. The complexity of construction environments renders traditional manual monitoring methods inefficient and costly. Therefore, developing efficient and automated helmet-wearing detection methods tailored to complex construction scenarios is imperative.

Object detection algorithms based on deep learning ~\cite{17, 18, 40, 41, 42} have become the main methods in the current field of object detection. These algorithms can generally be divided into two categories: two-stage methods and single-stage methods. Two-stage algorithms, including the R-CNN series (e.g., R-CNN, Fast R-CNN, Faster R-CNN, and Mask R-CNN), first generate region proposals to determine the locations that may contain objects, and then classify and further refine these proposals to improve detection accuracy. Although this methods can achieve high detection accuracy, it has a high computational complexity due to the need for additional region generation steps and multiple stages of calculation, resulting in relatively slow processing speed, especially in applications with high real-time requirements.

Compared with the two-stage method, single-stage algorithms such as SSD ~\cite{14} and YOLO series combines region proposal generation and classification operations into one step, avoiding the complex process of proposal generation and subsequent processing in the two-stage method, thereby achieving faster detection speed. Due to this simplified processing flow, the single-stage algorithm has a significant advantage in processing speed and is therefore widely used in real-time detection applications, especially in scenarios that require rapid response, such as construction sites. For example, YOLOv2 is optimized based on its original architecture by introducing dense blocks and lightweight MobileNet structures to enhance the ability to detect small objects, while minimizing resource consumption by reducing computational complexity, thereby achieving faster detection speed and higher detection accuracy ~\cite{15}.

The detection task studied in this paper presents three primary challenges compared to other object detection tasks. First, safety helmets are relatively small, resulting in fewer pixels and smaller feature representations extracted by the network compared to larger objects like pedestrians or vehicles. Second, in densely populated scenes, safety helmets often overlap, reducing the effective pixel area available for detection. Finally, construction sites and factories often feature background colors similar to those of the helmets, increasing the likelihood of missing detections.The motivation of this paper is to address the mentioned three challenges above, making the algorithm more suitable for the detection task studied in this paper.

In contrast to general object detection, which typically involves identifying objects with high variability in appearance and dynamic, cluttered backgrounds, safety helmet detection focuses on a specific object with relatively uniform shape and characteristics in controlled environments~\cite{55}. The primary challenge lies in accurately detecting small helmets across varying scales, angles, and occlusions caused by buildings, all within noisy construction site backgrounds~\cite{56}. To address these challenges, this paper proposes an enhanced YOLOv8 model that significantly improves real-time detection performance for safety helmet compliance in construction environments. Through targeted optimizations, our approach meets the critical requirements for effective deployment on edge devices in the field.
Our contributions are listed as follows:
Our contributions are summarized as follows:  
\begin{itemize}  
\item We introduce the Dynamic Attention Detection Head (DAHead) into the YOLOv8 model, enhancing its capability to detect small targets effectively.  

\item We replace the original Progressive Attention Feature Pyramid Network (PAFPN) with a Bi-directional Weighted Feature Pyramid Network (BWPPN), achieving more comprehensive fusion of input features across different scales.  

\item We substitute the CIoU loss with the Wise-IoU loss, improving convergence speed and overall model efficiency.  
\end{itemize}


\section{RELATED WORK}

\subsection{Object Detection Method}
The main goal of object detection is to automatically identify and locate objects of a specific category from images or videos. Unlike image classification tasks, object detection not only needs to determine whether there are objects of a certain category in the image, but also needs to determine the location and size of the object to generate a bounding box containing the object and classify the object. Traditional methods~\cite{36, 37, 38, 42}, such as those by Silva et al. leveraging HOG features and Fan et al. employing Kalman filters with Cam-shift algorithms, rely heavily on manual feature engineering and specific classifiers. While effective in certain scenarios, these approaches often suffer from limited generalization and high computational costs.
Diffusion models~\cite{61} have recently emerged as a promising paradigm in object detection, leveraging iterative denoising processes to refine bounding box predictions and enhance feature representation. 
With advancements in technology, deep learning-based methods have become dominant, typically categorized into two-stage and single-stage approaches. Two-stage methods, such as the R-CNN series, first generate region proposals, followed by classification and regression. In contrast, single-stage approaches, like YOLO, SSD, and RetinaNet, directly predict classes and bounding boxes on feature maps~\cite{57}. Although traditionally less accurate, recent innovations have enabled single-stage methods to surpass their two-stage counterparts in both speed and precision~\cite{58}. 
Among these, YOLOv8 represents the latest evolution of the YOLO series, incorporating optimized backbones and multi-scale feature fusion techniques to enhance detection accuracy and real-time performance while maintaining a lightweight and flexible design.

\subsection{Attention Mechanism}
The attention mechanism~\cite{41, 52, 53, 54} has shown great potential in object detection by enabling models to focus on relevant areas and optimize computational resource allocation. Notably, Detection Transformer (DETR), introduced by Carion et al., enhances object relationship modeling within the global image context, outperforming traditional two-stage detection methods. However, DETR’s high computational requirements present challenges for processing large images and localizing small objects. More recent feature pyramid attention mechanisms, as proposed by Zhang et al. and Li et al., have advanced multi-scale feature representation with minimal computational overhead, driving progress in real-time safety helmet detection. 

For example, MFC~\cite{54} can enhance dense target features in the frequency domain by introducing a frequency domain filtering module, thereby improving the model's ability to identify targets in complex backgrounds. PBSL~\cite{43} uses a multimodal alignment method to strengthen the connection between related features while effectively suppressing irrelevant information, thereby improving the quality of feature representation. CFIL~\cite{52} further combines frequency domain feature extraction and feature interaction modules, which can not only perform more refined feature extraction in the frequency domain, but also promote the representation of significant features through interaction between features. Based on these technical advances, we integrate the dynamic DAHead module into our target detection model in this study. This module uses an adaptive attention mechanism to optimize feature aggregation between pyramid layers, thereby achieving more accurate multi-scale feature fusion. The feature aggregation method can effectively enhance the complementarity of feature information at different scales, making the model more robust when dealing with multi-scale targets. In addition, by dynamically adjusting the attention weight, the DAHead module can adaptively adjust the influence of features at each level, further enhancing the model's ability to detect targets. Compared with the traditional YOLOv8 model, our method shows significant performance improvement on multiple datasets, especially in the detection accuracy of small objects and complex scenes.

\begin{figure*}[t]
\centering
\includegraphics[scale=0.55]{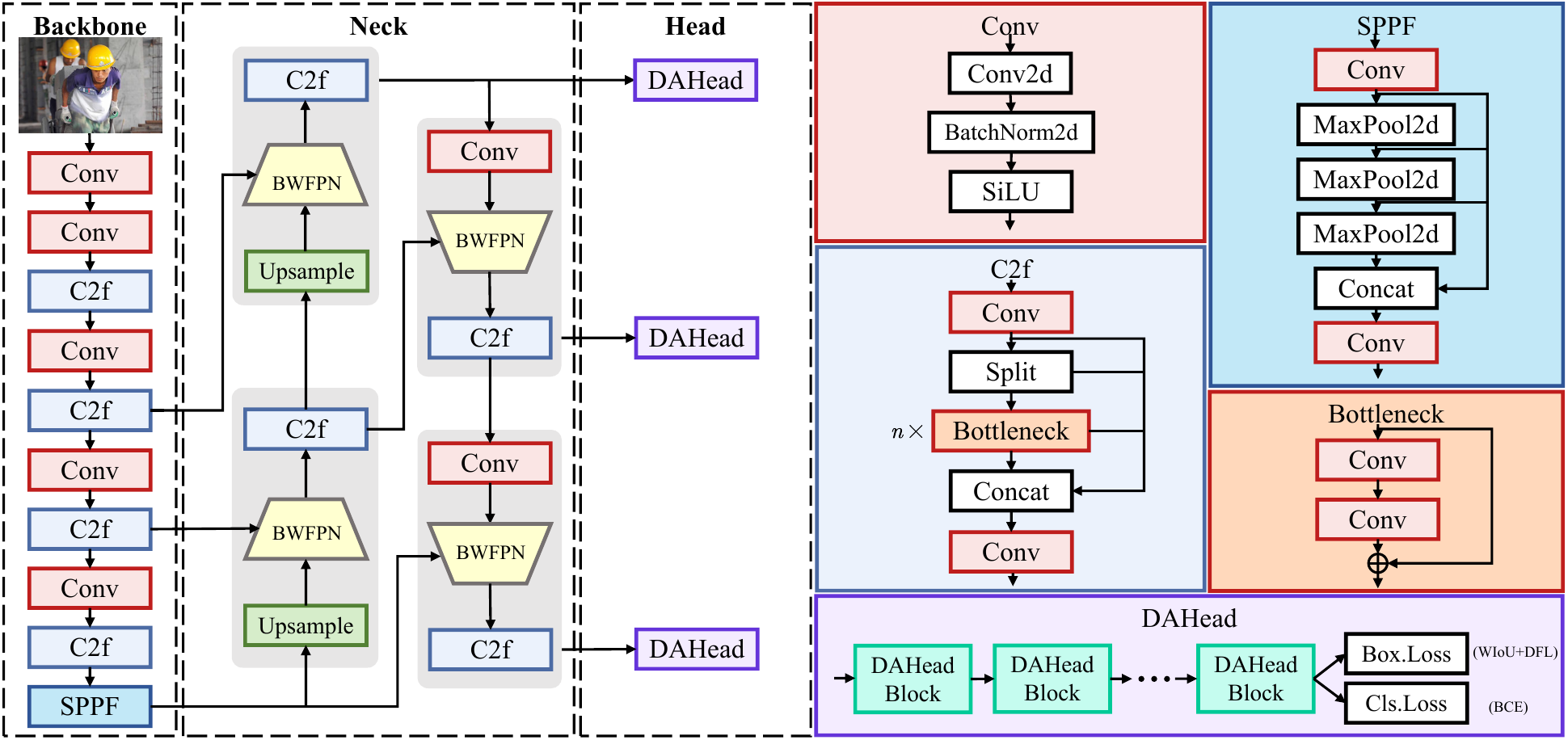}
\caption{DABFNet model framework.The Backbone is responsible for feature extraction through successive convolutional and C2f layers, culminating in a Spatial Pyramid Pooling Fast (SPPF) module that consolidates multi-scale information. The Neck integrates features using the BWFPN and upsampling modules. It connects multiple C2f modules and BWFPN layers to enhance feature representation and improve the network’s detection capability. The Head includes a series of DAHead Blocks that predict bounding boxes and classify objects.}
\label{fig1}
\vspace{-3mm}
\end{figure*}

\section{OUR PROPOSED METHOD}
\subsection{Improvement Framework Overview}
To improve the detection performance of YOLOv8 in our task, we propose the \textbf{D}ynamic \textbf{A}ttention and \textbf{B}i-directional \textbf{F}usion \textbf{Net}work (DABFNet) for
safety helmet wearing detection in this section. First, the Dynamic Attention Mechanism Detection Head (DAHead) is introduced into YOLOv8 model to improve the model’s ability to detect small
targets. Subsequently, we replace the original Progressive Attention Feature Pyramid Network (PAFPN) structure with Bi-directional Weighted Feature Pyramid Network (BWPPN) to improve the feature fusion efficiency. Finally, the original CIOU loss is replaced by Wise-IoU to guide the network to train more efficiently. The framework of DABFNet is shown in Fig.~\ref{fig1}.

\subsection{Dynamic Attention Detection Head}
DAHead uses an attention to unify different object detection heads. 
Given a feature pyramid-shaped feature output \(F_{in}= \left \{F_{i} \right \} _{i}^{L} \), \(L\) is the number of layers of the pyramid. 

Because the features of the three levels come from different sizes, it has multi-scale detection capabilities. The features of these three levels are fused to obtain \(f\in R^{L \times H\times W\times C}\) , which is further transformed into \(R^{L\times A\times C}\),  using \(S=H\times W\). Then, by paying attention to \(L\), \(S\) and \(C\) respectively, we can obtain three kinds of perception abilities.

We first apply self-attention on a feature layer as follows:
\begin{equation}
W(F)=\pi(F) \cdot F.
\end{equation}
where \( \pi(\cdot) \) is an attention mechanism.
Divide attention into three dimensions, and each attention focuses on only one dimension. The formula obtained is:
\begin{equation}
\mathrm{W}(\mathrm{F})=\pi_C\left(\pi_S\left(\pi_L(F) \cdot F\right) \cdot F\right) \cdot F.
\end{equation}
where C, S, and L represent applying attention on the C, S, and L dimensions respectively.
Scale-awareness attention gives different weights to feature layers, so that the model can adaptively fuse features based on the importance of the level:

\begin{equation}
\pi_L(F) \cdot F=\sigma\left(\mathrm{f}\left(\frac{1}{S C} \sum_{S, C} F\right)\right) \cdot \mathrm{F}.
\end{equation}
where \( \frac{1}{S C}\sum_{S, C} F \) is an average pool,  \( f(\cdot) \) is a convolution function, and \(\sigma(\mathrm{x})\) is a hard-sigmoid activation function.
Spatial-awareness attention focuses on the ability to distinguish different spatial locations. Due to the high dimensionality of S, deformation
convolution is used to sparse the attention, and then features are integrated across scales:
\begin{equation}
\pi_S(F) \cdot F=\frac{1}{L} \sum_{l=1}^L \sum_{k=1}^K \omega_{l, k} \cdot F\left(l ; p_k+\Delta_{p_k} ; c\right) \cdot \Delta_{m_k}.
\end{equation}
where K is the number of sparsely sampled locations, \( p_{k} + \Delta_{p_{k}} \) is a location shifted by the spatial offset \(\Delta_{p_{k}}\), and \(\Delta_{p_{k}}\) is a scalar at location \(p_k\).

Task-aware attention is designed to promote the generalization of joint learning and target expression ability. It can dynamically switch feature channels to assist different tasks:

\begin{equation}
\small
\pi_C(F) \cdot F = \max \left(\alpha^1(\mathrm{F}) \cdot F_c + \beta^1(\mathrm{F}), \alpha^2(\mathrm{F}) \cdot F_c + \beta^2(\mathrm{F})\right).
\end{equation}
where \(\left[\alpha^1, \alpha^2, \beta^1, \beta^2\right]^T=\theta(\cdot) \) is a hyperparameter used to control the activation threshold, and \( \theta(\cdot)\) is similar to Dynamic Rectified Linear Unit (DyReLU).

\begin{figure}[t]
\centering
\includegraphics[scale=0.4]{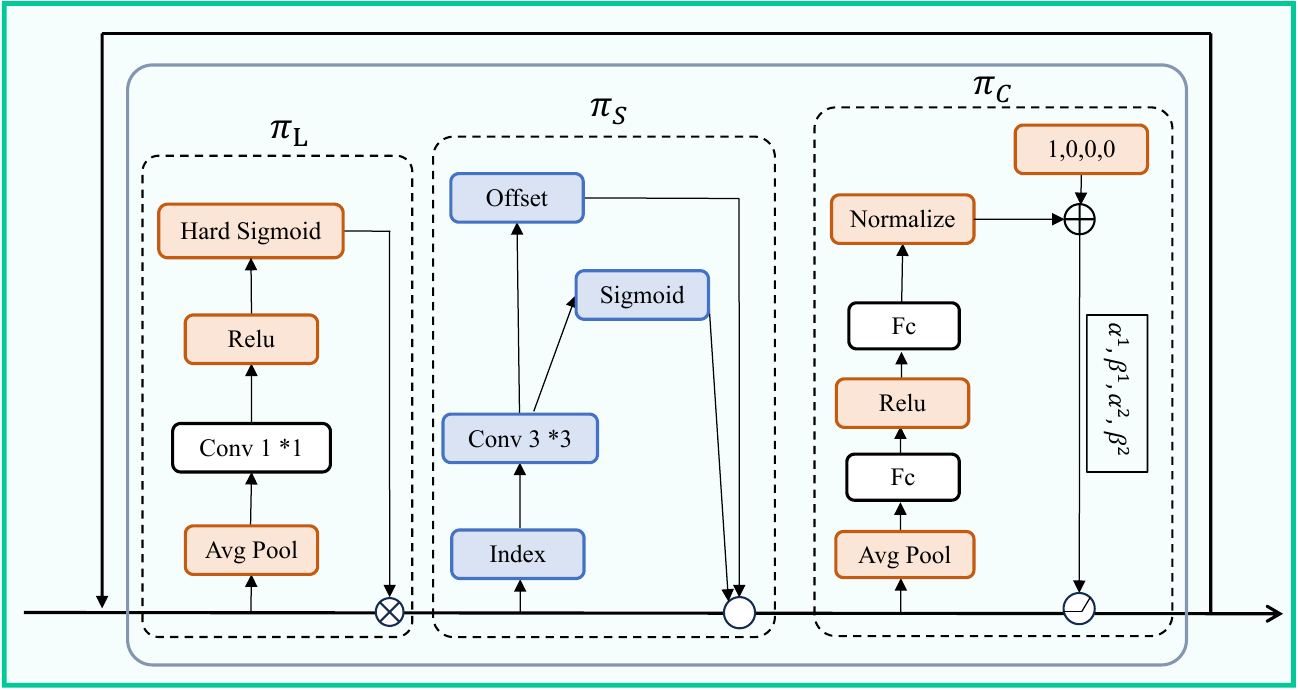}
\caption{Dynamic Attention Detection Head Block. The Dynamic Attention Head Block comprises three components: \(\pi_{L}\), responsible for local feature processing; \(\pi_{S}\), which refines spatial features through convolution and offset adjustments; and \(\pi_{C}\) focused on channel-wise feature modulation.}
\label{fig2}
\vspace{-3mm}
\end{figure}

As shown in Fig. ~\ref{fig2}, the implementation process of each attention module is described. We designed three different types of attention modules to optimize different dimensions of the feature tensor F. Specifically, the scale-aware attention (\(\pi\)L) focuses on weighting features of different scales, extracts the global information of features using average pooling operations, and then further processes the features through convolutional layers. Next, the ReLU activation function is used for nonlinear transformation, and finally the hard sigmoid function is used to generate scale-aware weight values to select the most representative parts among features of different scales. The core idea of the spatial-aware attention (\(\pi\)S) is to focus on spatially important areas by learning the spatial structure of the feature map. The task-aware attention (\(\pi\)C) linearly transforms the features through a fully connected layer to obtain the weight coefficients of each channel, and enhances the nonlinear expression through the ReLU activation function. Finally, the module performs normalization to ensure that the obtained weights remain stable during training.

\subsection{Bi-directional Weighted Feature Pyramid Network}
This section shows the improvement of the feature fusion method proposed to improve the detection effect for small targets and occluded targets.

\begin{figure}[t]

    \subfigure[FPN]{
        \begin{minipage}{0.14\linewidth}
          \includegraphics[width=1\linewidth]{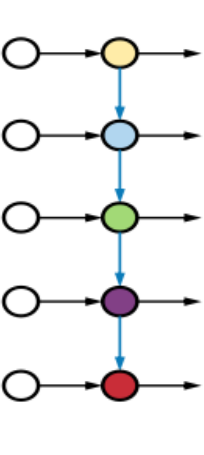} 
          \vspace{-13pt} 
         \end{minipage}
         }
    \subfigure[PANet]{
        \begin{minipage}{0.2\linewidth}
          \includegraphics[width=1\linewidth]{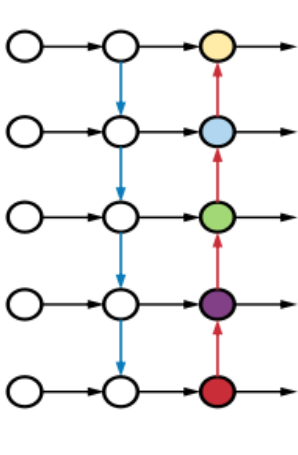}
          \vspace{-12pt} 
         \end{minipage}         
        }
    \subfigure[NAS-FPN]{
        \begin{minipage}{0.24\linewidth}
          \includegraphics[width=1\linewidth]{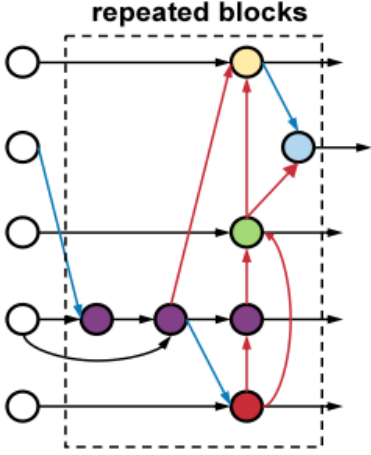} 
          \vspace{-11pt} 
         \end{minipage}      
        }
    \subfigure[ours]{
        \begin{minipage}{0.24\linewidth}
          \includegraphics[width=1\linewidth]{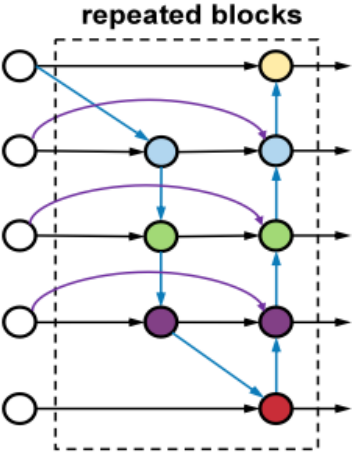} 
         \end{minipage}         
        }
\centering
\caption{Comparison of different feature fusion methods. (a) FPN, which uses a simple top-down pathway for multi-scale feature fusion; (b) PANet, which adds a bottom-up pathway to enhance information flow; (c) NAS-FPN, which leverages neural architecture search to create an optimized, multi-level feature fusion structure with repeated blocks;  (d) our BWFPN, which incorporates bidirectional connections and repeated blocks for efficient and adaptive feature fusion across different scales.}
\label{fig3}
\vspace{-3mm}
\end{figure}

In the traditional feature fusion, suppose there is a column of multi-scale features 

\begin{equation}
\overset{\rightarrow}{P}^{i n} = \left(P_{l_{1}}^{i n} , P_{l_{2}}^{i n} , \ldots\right).
\end{equation}
Where \(P_{l_{i}}^{i n}\) represents the feature of level \(l_{i}\) The task of feature fusion is to find a mapping function that satisfies: \(\vec{P}^{out}=f{\left(\vec{P}^{in}\right)}\), which fully integrate different features into new features. The traditional FPN structure is shown in Fig.~\ref{fig3}(a). It uses 3 to 7 levels of input features, and the resolution is halved as the level increases. 
The traditional top-down aggregation method of FPN suffers from one-way information flow limitations. To address this, PANet introduced bidirectional feature fusion, incorporating a bottom-up path for aggregation, as illustrated in Fig.~\ref{fig3}(b). Building on this, NAS-FPN (Fig.~\ref{fig3}(c)) leveraged neural structure search to design a cross-scale feature network topology. BWFPN further enhances cross-scale connections with three key optimizations: it removes nodes with a single input edge, improving network efficiency; it adds an extra path from the original input to the output node, allowing for better feature fusion; and it treats each bidirectional fusion network as a stackable layer for more thorough feature integration. The BWFPN structure is depicted in Fig.~\ref{fig3}(d). Recognizing that input features from different scales have varying resolutions and contributions to the output, BWFPN assigns additional weights to each input.
\begin{equation}
O=\sum_i\frac{w_i}{\epsilon+\sum_jw_j}\cdot I_i.
\end{equation}
\(\epsilon\) is a sufficiently small value to avoid the instability of numerical calculations, and the final weight is mapped to a value of \([0,1]\). Compared with the original YOLOv8 structure, BWFPN can more fully fuse input features of different scales.

\subsection{Optimization Objectives}

This paper studies replacing the CIoU loss function with Wise-IoU (WIoU). WIoU dynamically calculates the IoU loss in the category prediction loss as follows:

\begin{equation}\begin{aligned}
L_{WIoU}&=R_{WIoU}\cdot L_{IoU}\\R_{WIoU}&=\exp\left(\frac{\left(x-x_{gt}\right)^{2}+\left(y-y_{gt}\right)^{2}}{\left(W_{g}{}^{2}+H_{g}{}^{2}\right)^{*}}\right).
\end{aligned}\end{equation}
The WIoU loss function has the characteristics of dynamic non-monotonicity, uses outlier points instead of naive IoU to evaluate the similarity between anchor boxes and annotation boxes, and provides a wise gradient gain distribution strategy. The design of this structure can improve the accuracy of model detection and accelerate the convergence of the model.

\section{EXPERIMENT}
\subsection{\textbf{Dataset And Hyper-parameters}}


\begin{figure}[t]
\centering
\subfigure[The instance counts of each label.]{
    \includegraphics[width=0.45\linewidth]{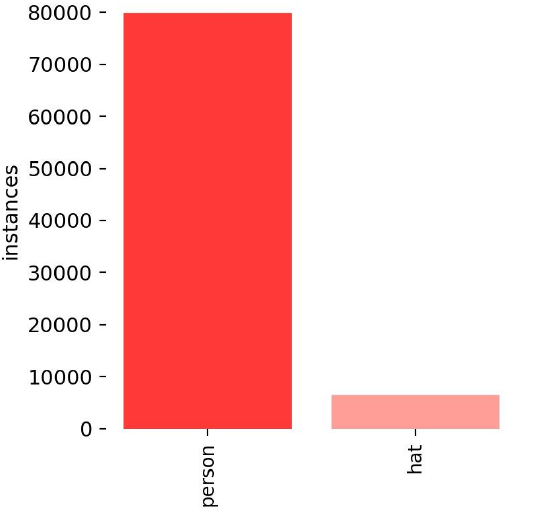}
}
\subfigure[The bounding box distribution.]{
    \includegraphics[width=0.45\linewidth]{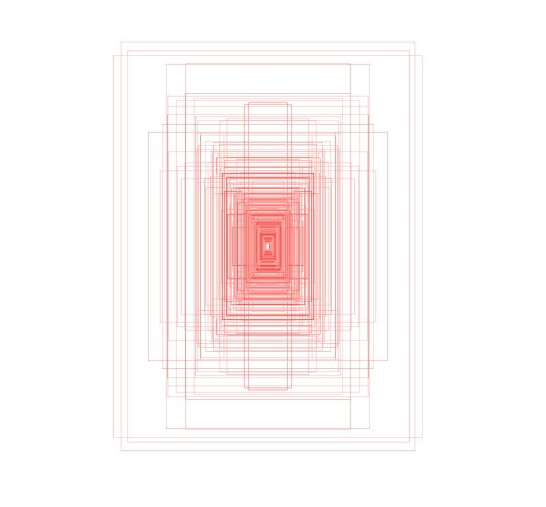}
}
\subfigure[Spatial distribution of bounding boxes.]{
    \includegraphics[width=0.45\linewidth]{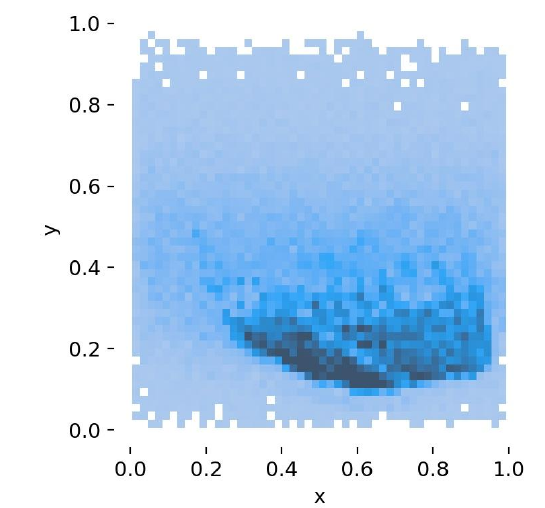}
}
\subfigure[The relationship between bounding box width
and height.]{
    \includegraphics[width=0.45\linewidth]{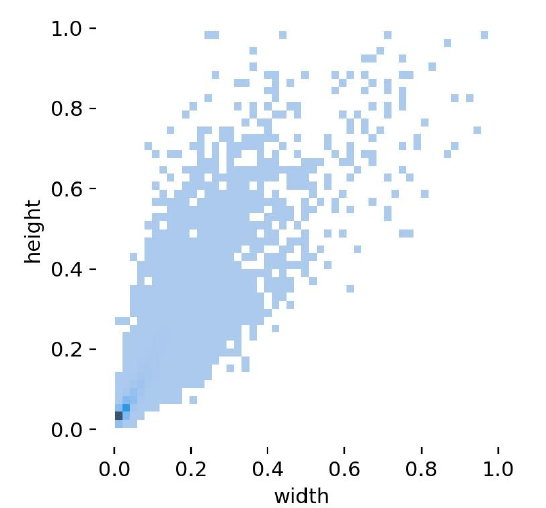}
}

\caption{Dataset label distribution. Fig.(a) shows the instance counts of each label, with "person" significantly outnumbering "hat"; Fig.(b) presents the bounding box distribution, revealing the common positions and sizes of annotated objects; Fig.(c) indicates the spatial distribution of bounding boxes on the x and y axes; Fig.(d) displays the relationship between bounding box width and height, suggesting variations in object sizes within the dataset.}
\label{fig4}
\vspace{-3mm}
\end{figure}

This paper uses the Safety Helmet Wearing Detection (SHWD) dataset  which contains a total of 7584 images, including 5450 training sets, 606 validation sets, and 1528 test sets. The image categories include people wearing helmets (hat) and people not wearing helmets (person). The image scenes are mostly construction sites, and the images are downloaded from the Internet and annotated. 
The experimental setups are shown in Tables~\ref{tab:1}, and \ref{tab:2}.

\begin{table}[t]
\centering
\caption{Training environment and hardware platform parameters table.}
\label{tab:1} 
\begin{tabular}{ >{\centering\arraybackslash}p{4cm}  >{\centering\arraybackslash}p{4cm} }
\hline
\textbf{Parameters}        & \textbf{Configuration}                                                                            \\ \hline
CPU                        & i7-8565U                                                                                          \\
GPU                        & NVIDIA GeForce RTX 4090D                                                                          \\
memory size                & 24G                                                                                               \\
Operating systems          & Unbantu20.04                                                                                      \\
Python          &  3.8.10                                                                                    \\
Pytorch          & 1.9.0                                                                                      \\
Cuda          & 11.1                                                                                      \\
Cudnn & \begin{tabular}[c]{@{}c@{}}11.4\end{tabular} \\ \hline
\end{tabular}
\end{table}

\begin{table}[t]
\centering
\caption{Some key parameters set during model training.}
\label{tab:2} 
\begin{tabular}{ >{\centering\arraybackslash}p{4cm}  >{\centering\arraybackslash}p{4cm} }
\hline
\textbf{Parameters}       & \textbf{Setup}     \\ \hline
Batch size       & 16        \\
Epochs           & 200       \\
Optimizer        & SGD       \\
Learning rate    & 0.01      \\
Momentum         & 0.9       \\
Input image size & 640 × 640 \\
Weight decay     & 0.0005    \\ \hline
\end{tabular}
\end{table}

 \subsection{\textbf{Model Evaluation Metrics}}
 This study evaluates model performance using these indicators: precision, recall, mean average precision (mAP) and GFLOPs. Precision measures the accuracy of positive predictions for each class, while recall assesses the model's sensitivity to actual positives. Average Precision (AP) evaluates detection performance across thresholds, and mAP provides an average performance measure across all classes, using mAP@0.5 and mAP@[.5:.95] for evaluation.GFLOPs indicate model size and computational efficiency.

\begin{figure}[t]
\centering
\includegraphics[scale=0.4]{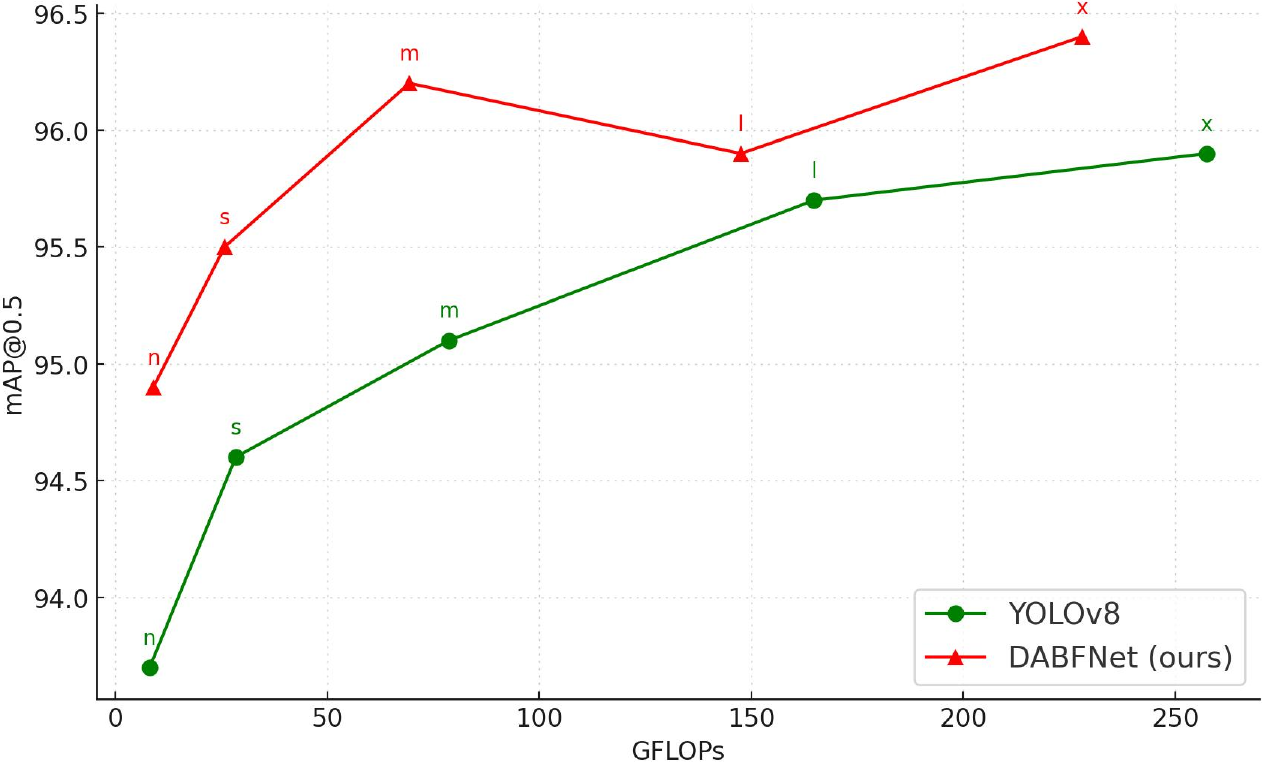}
\caption{Comparison of models of different sizes. This figure presents a comparison of models of different sizes based on mAP@0.5 and GFLOPs. The DABFNet model (in red) consistently outperforms YOLOv8 (in green) across all sizes, achieving higher mAP@0.5 values at comparable computational costs.}
\label{fig7}
\vspace{-3mm}
\end{figure}

\begin{figure}[t]
\centering
\includegraphics[scale=0.33]{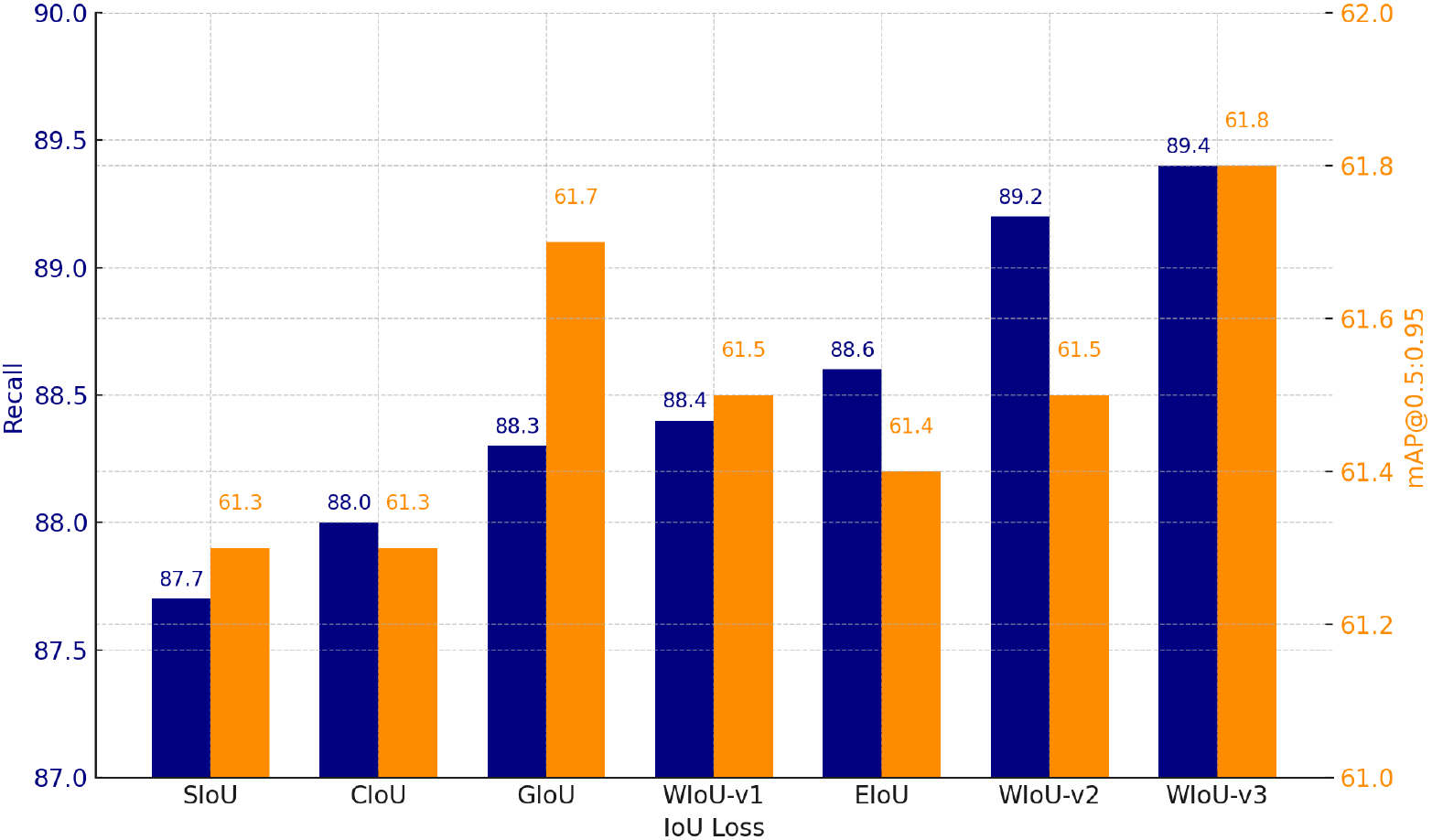}
\caption{Loss function comparison experiment. This figure displays the results of a loss function comparison experiment, with recall (in blue) and mAP@[0.5:0.95] (in orange) metrics across various loss functions. The comparison includes IoU-based losses (SIoU\cite{63}, CIoU\cite{34}, GIoU\cite{65}) and weighted IoU variants (WIoU-v1\cite{67}, WIoU-v2\cite{67}, WIoU-v3), as well as EIoU\cite{66}. The WIoU-v3 achieves the highest recall and mAP values, indicating its effectiveness in optimizing model performance. This suggests that WIoU-v3 provides a more robust training objective, enhancing both precision and recall in object detection tasks compared to other loss functions.}
\label{fig8}
\vspace{-3mm}
\end{figure}

\begin{figure}[t]
\centering
\includegraphics[scale=0.35]{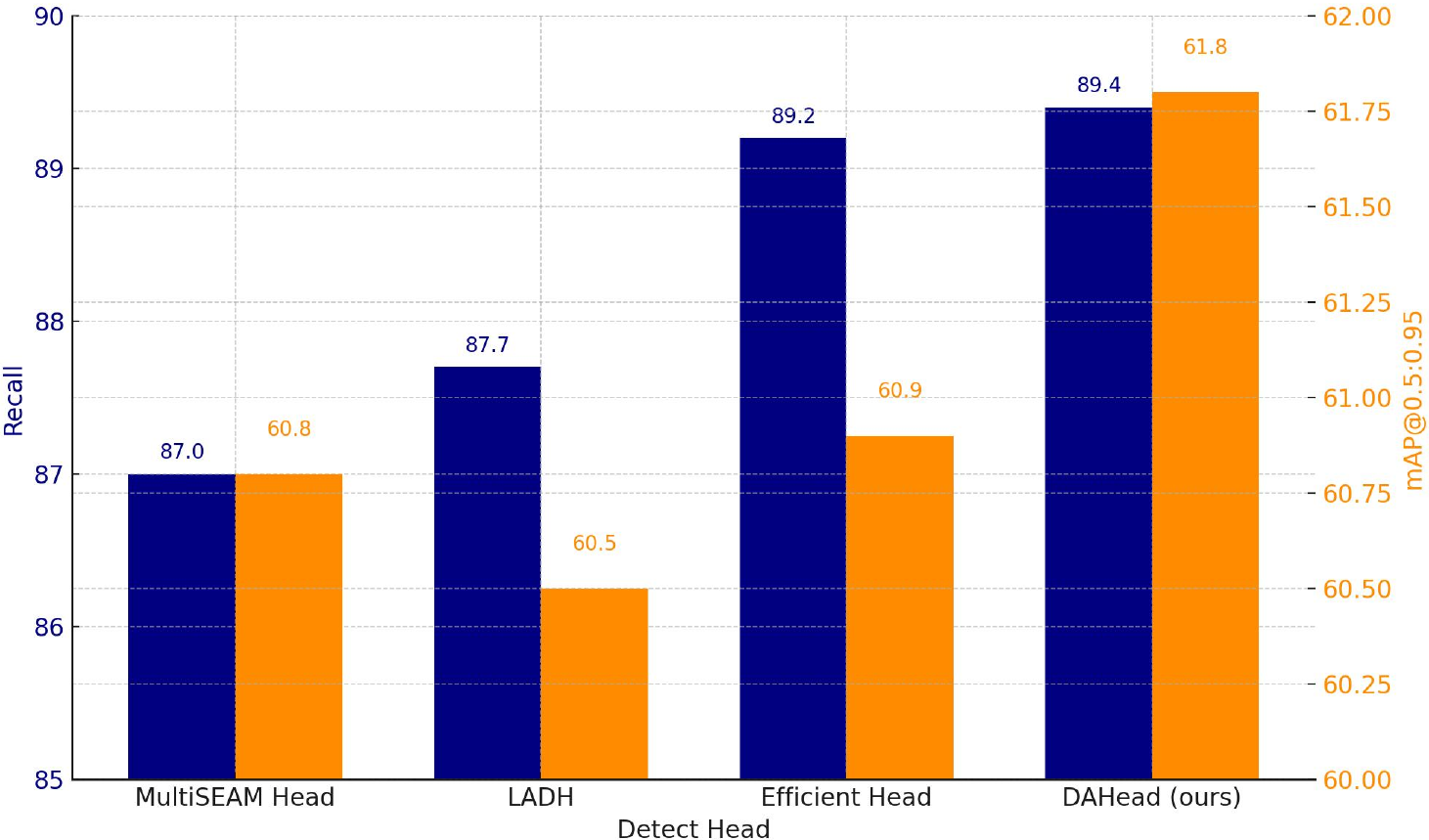}
\caption{Detection head comparison experiment.This figure shows the recall (in blue) and mAP@[0.5:0.95] (in orange) values of models under different detection heads (MultiSEAM Head\cite{69}, LADH\cite{70}, Efficient Head\cite{28} and DAHead). It is easy to observe from the figure that our model's dynamic attention mechanism detection head has the highest recall rate and mAP@[0.5:0.95] value, making it more suitable as the detection head of our model.}
\label{fig9}
\vspace{-3mm}
\end{figure}

\subsection{\textbf{Baseline Comparison Experiment}}
\subsubsection{Baseline Model}
YOLOv3-tiny is a simplified version of YOLOv3, which 
 can meet the needs of the detection scenario in this article. Therefore, YOLOv3-tiny is selected as one of the baseline models.
The network structure of YOLOv5 can be divided into a backbone network, a neck network, and a head for detection. The neck network uses a combination of FPN and PA network. The backbone network uses a bottom-up path to extract features from the original image. In the detection stage, shallow feature maps that fuse the semantic information are used to identify small targets, and then deeper feature maps are used to identify large targets. YOLOv5 has been widely used in the industry due to its high accuracy and fast detection speed.
YOLOv5-p6 This is the fifth update of YOLOv5. Compared with the three output layers of YOLOv5-P5, namely P3, P4, and P5, the output layer of the YOLOv5-P6 model has an additional P6, which is mainly used to detect large targets at high resolution.
YOLOv6 focuses on both detection accuracy and inference efficiency, and absorbs a lot of ideas from different network designs, training strategies, testing techniques, quantization, and optimization methods. Compared with previous versions, YOLOv6 customizes models of different sizes for different scenarios. Small models are characterized by ordinary single-path trunks, while large models are built on efficient multi-branch blocks. A self-distillation strategy is added to perform classification and regression tasks at the same time, and various advanced tricks are integrated, such as label assignment detection technology, loss function, and data enhancement technology. With the help of re-optimizers and channel distillation, the quantitative detection scheme is reformed to obtain a better detector.

\subsubsection{Experimental Results And Discussion}
To verify the effectiveness and reliability of the algorithm proposed, we compare the algorithm proposed with the current baselines under the same hardware and software environment. For the fairness of the experiment, the model size is uniformly set to the default "n". From the results in Table~\ref{tab:3}, we can see that on different categories of data on the safety helmet wearing dataset, the average precision of DABFNet is on par with the best performing baseline algorithm YOLOv5-p6, the average recall is 1.3\% higher than the best performing baseline algorithm YOLOv5, the average mAP@0.5 is 1.2\% higher than the best performing baseline algorithm YOLOv8, and the average mAP@[0.5:0.95] is 1.7\% higher than the best performing baseline algorithm YOLOv8.
In addition, the Recall and mAP@0.5 of DABFNet on the Hat category data are 1.3\% and 1.1\% higher than the best performing baseline algorithm, respectively, while the  Recall and mAP@0.5, on the Person category data are  1.2\% and 1.1\% higher than the best performing baseline algorithm, respectively.

\subsection{\textbf{Ablation Experiment}}
To verify the effectiveness of the improvements made and quantify the degree of improvement of each improvement on model performance, this paper conducted an ablation experiment, and 
indicates that this improvement strategy is used. The experimental results are shown in Table~\ref{tab:9}. Compared with the YOLOv8 baseline model, adding the DAHead module alone increased the precision by 0.3\%, the recall by 1.0\%, the mAP@0.5 by 0.4\%, and the mAP@[.5:.95] by 0.4\%, verifying that the DAHead improves the model's accuracy and speed. By improving the feature fusion network, the recall rate was improved by 0.6\%, mAP@0.5 was improved by 0.1\%, and mAP@[.5:.95] was improved by 0.2\%, which verified that the BWFPN feature fusion layer improves the algorithm performance. By adding the DAHead and the BWFPN at the same time, the model effect is better than using either one alone. After using the WIoU-v3 loss function, the recall rate increased by 1.7\%, and the mAP@0.5 increased by 0.1\%. Compared with before the improvement, the WIoU-v3 loss function further improved the model performance. After using the three strategies of DAHead, BWFPN and WIoU-v3 at the same time, the model's Recall, mAP@0.5 and mAP@[0.5:0.95] indicators are all improved. Therefore, this ablation experiment proves the effectiveness of the three improved methods proposed individually and in relation to each other.

 \subsection{\textbf{Module Comparison Experiment}}
  To verify the effectiveness of the improved loss function, this paper compares the impact of several different loss functions on the model performance. The experimental results are shown in Fig.~\ref{fig8}. Compared with CIoU used by YOLOv8, the effects of EIoU and DIoU have obviously declined. However, after replacing CIoU with WIoU-v3, the recall rate of the model on the validation set increased by 1.4\%,  mAP@[0.5:0.95] increased by 0.3\%, which proves the rationality and effectiveness of our loss function replacement. To verify the effectiveness of the DAHead proposed in this paper, this paper compares the impact of several different detection heads on model performance. The experimental results are shown in Fig.~\ref{fig9}. For the fairness of the experiment, the number of detection heads is the same. Comparison shows that the DAHead has achieved the best results in terms of recall rate and mAP@[0.5:0.95] compared with others, which verifies the effectiveness of the attention mechanism.
%

\begin{figure}[t]
\centering
\subfigure[YOLOv3-tiny]{
    \begin{minipage}{0.27\linewidth}
      \centering\includegraphics[width=1\linewidth]{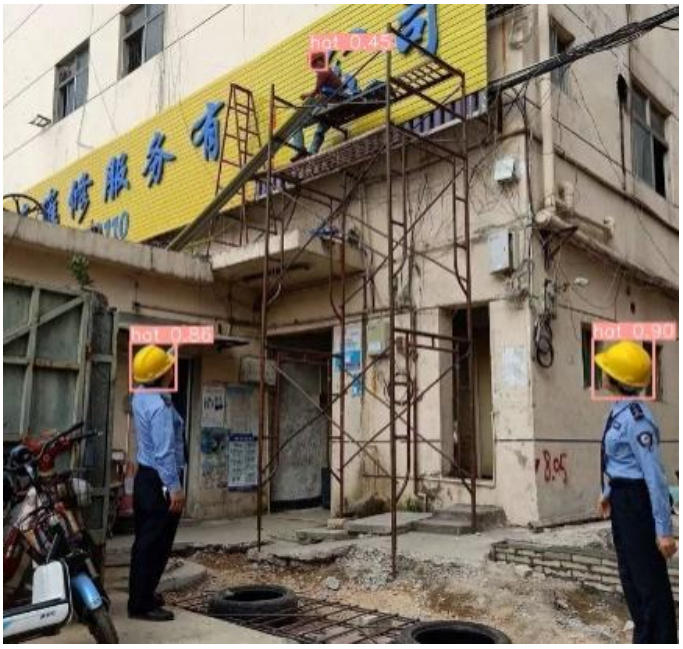}\vspace{0.04cm}
      \centering\includegraphics[width=1\linewidth]{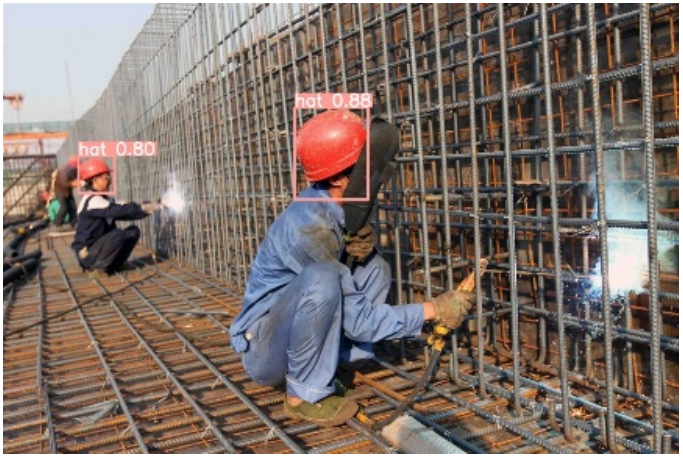}\vspace{0.04cm}
      \centering\includegraphics[width=1\linewidth]{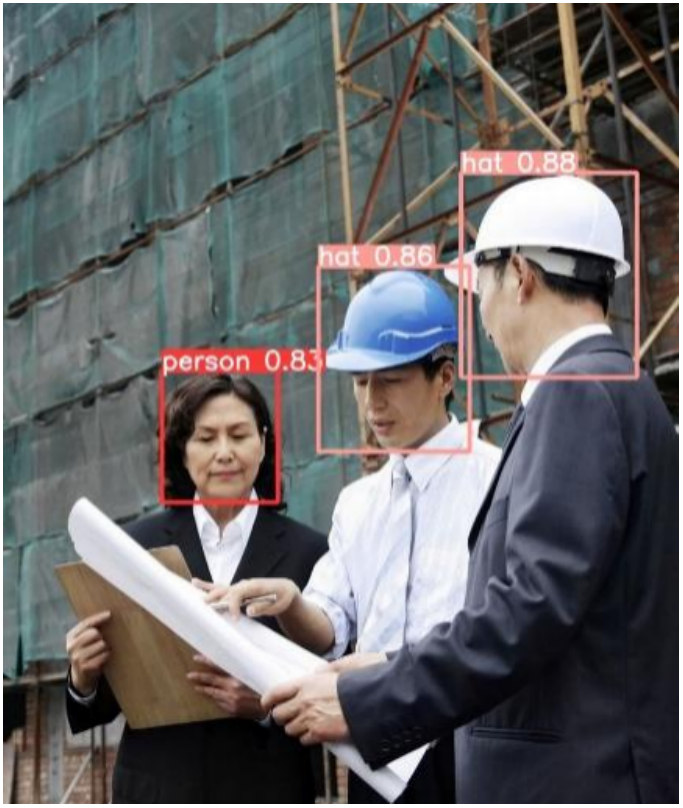}
    \end{minipage}
}
\subfigure[YOLOv8]{
    \begin{minipage}{0.27\linewidth}
      \centering\includegraphics[width=1\linewidth]{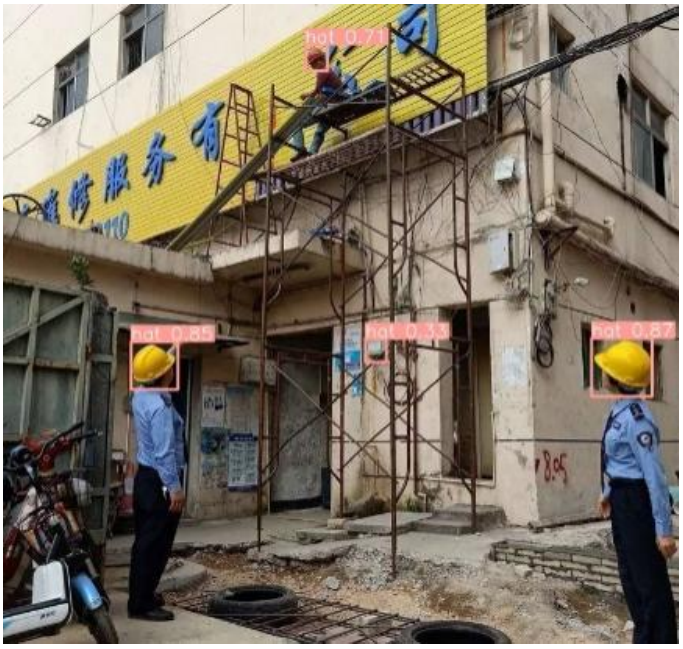}\vspace{0.04cm}
      \centering\includegraphics[width=1\linewidth]{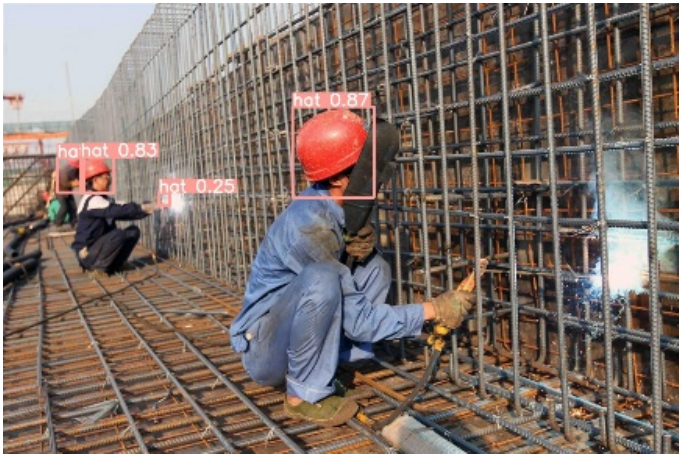}\vspace{0.04cm}
      \centering\includegraphics[width=1\linewidth]{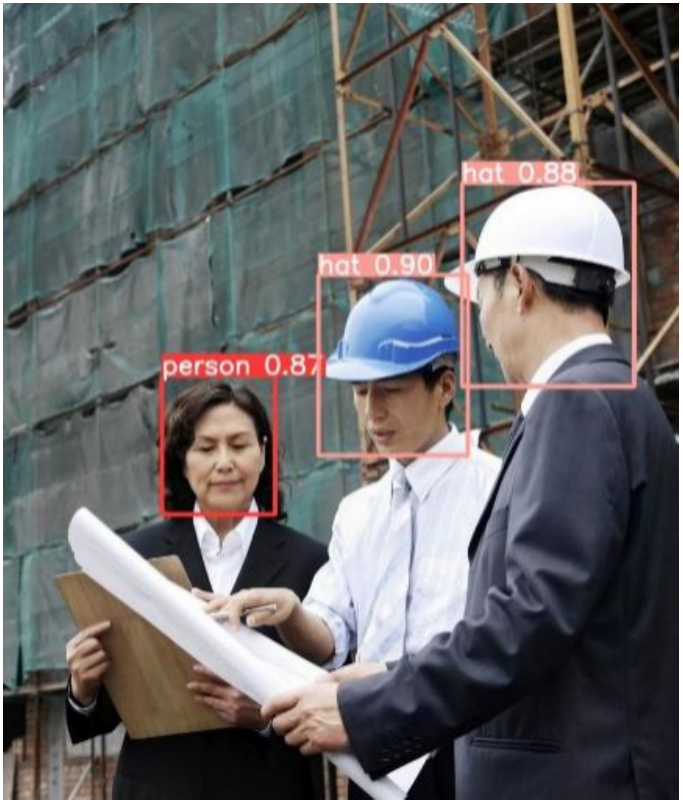}
    \end{minipage}         
}
\subfigure[DABFNet]{
    \begin{minipage}{0.27\linewidth}
      \centering\includegraphics[width=1\linewidth]{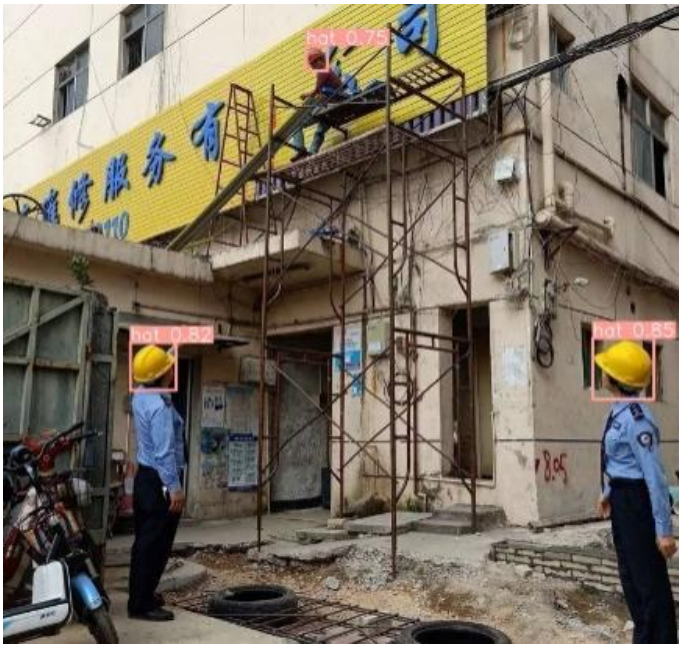}\vspace{0.04cm}
      \centering\includegraphics[width=1\linewidth]{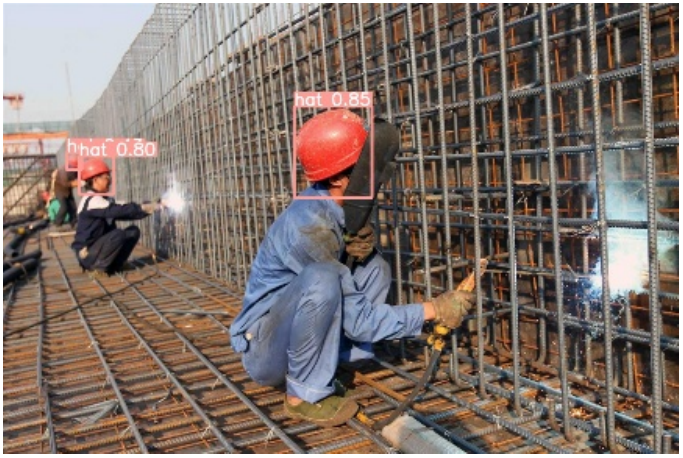}\vspace{0.04cm}
      \centering\includegraphics[width=1\linewidth]{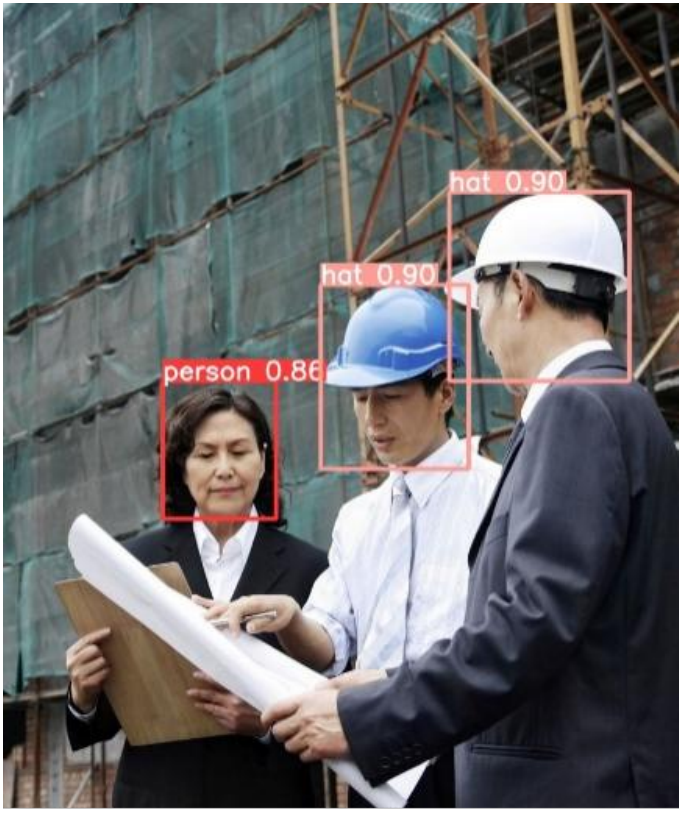}
    \end{minipage}         
}         
\caption{Comparison of detection results without heatmaps. YOLOv3-tiny, YOLOv8, and DABFNet detection results on different scenes.}
\label{fig5}
\end{figure}

\begin{figure}[t]
\centering
\subfigure[YOLOv3-tiny]{
    \begin{minipage}{0.27\linewidth}
      \centering\includegraphics[width=1\linewidth]{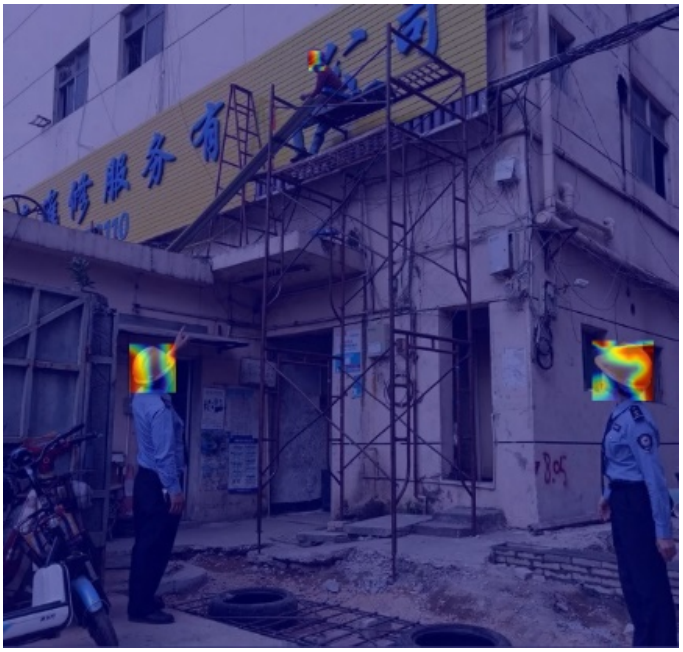}\vspace{0.04cm}
      \centering\includegraphics[width=1\linewidth]{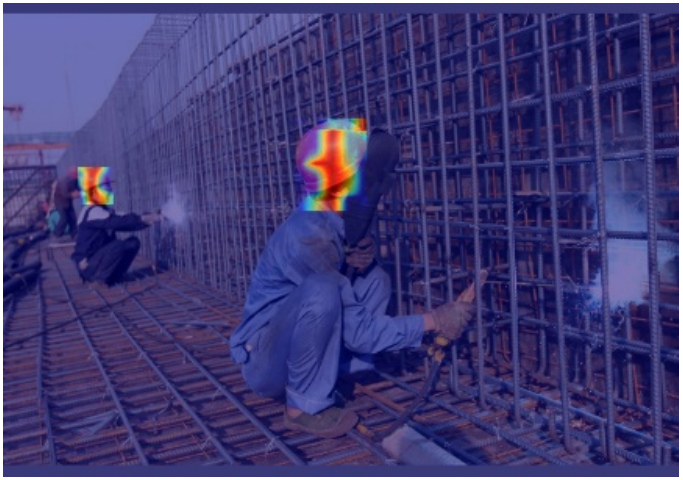}\vspace{0.04cm}
      \centering\includegraphics[width=1\linewidth]{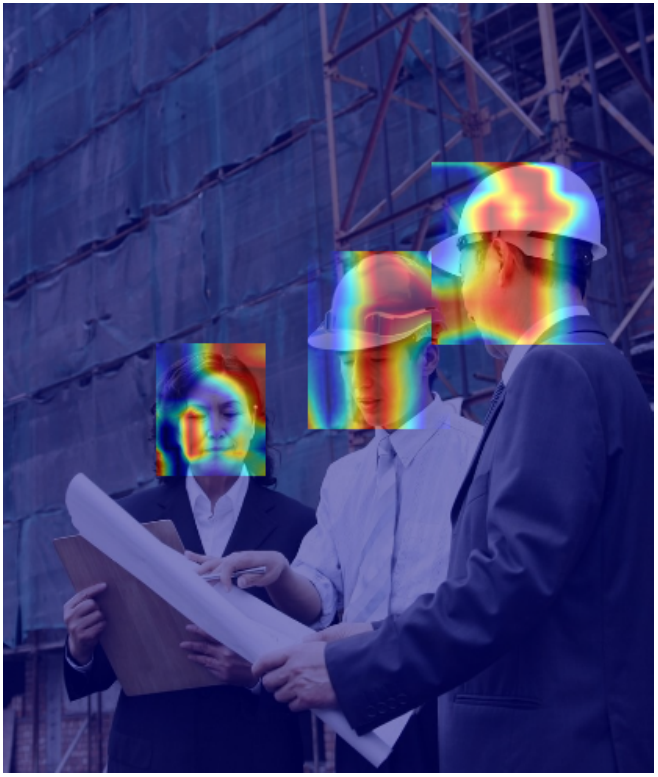}
    \end{minipage}
}
\subfigure[YOLOv8]{
    \begin{minipage}{0.27\linewidth}
      \centering\includegraphics[width=1\linewidth]{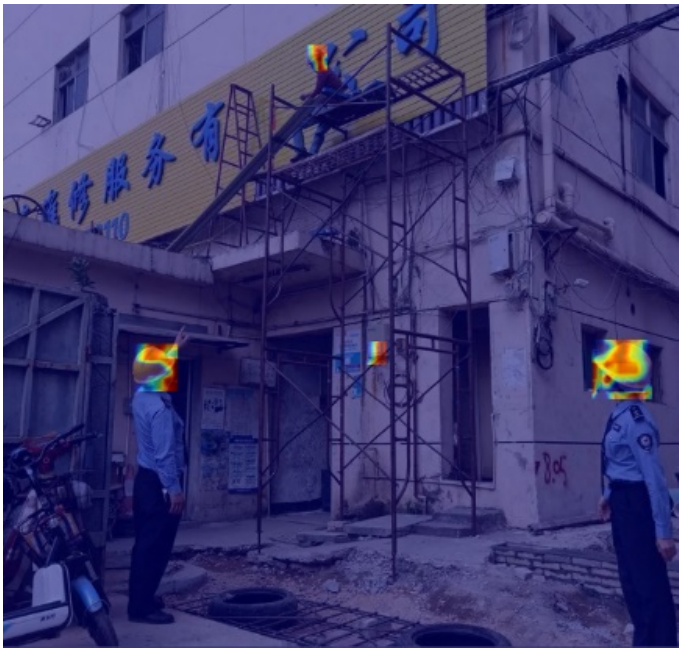}\vspace{0.04cm}
      \centering\includegraphics[width=1\linewidth]{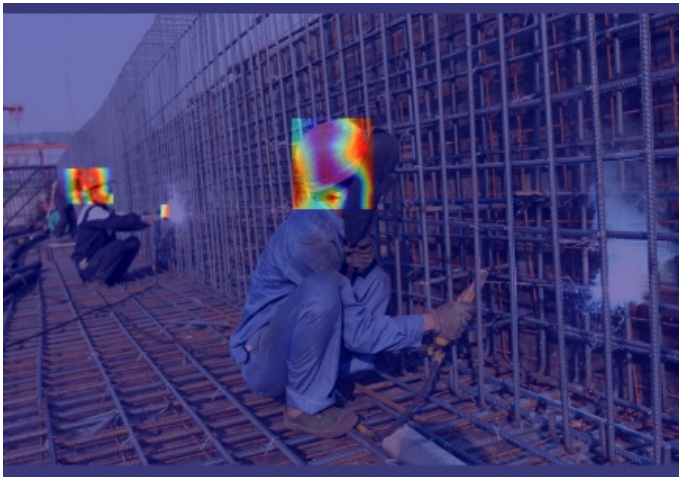}\vspace{0.04cm}
      \centering\includegraphics[width=1\linewidth]{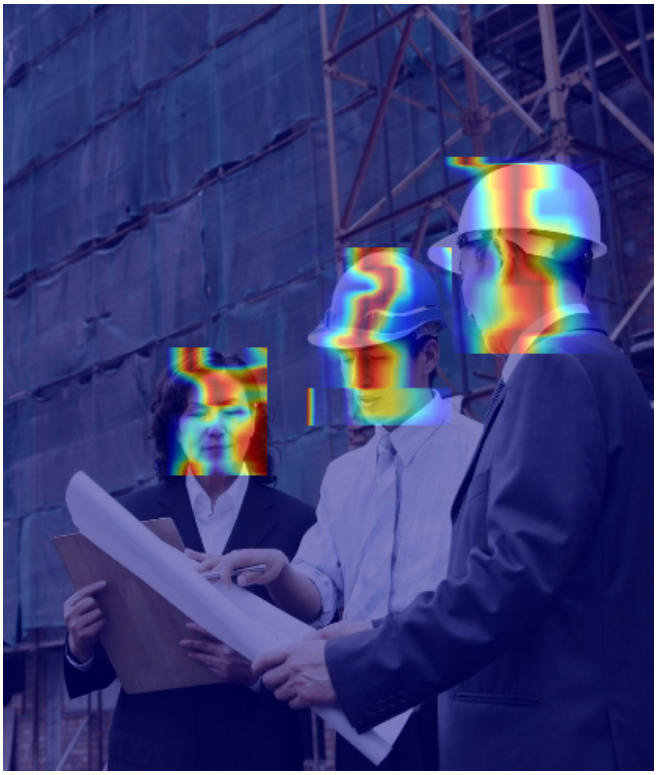}
    \end{minipage}         
}
\subfigure[DABFNet]{
    \begin{minipage}{0.27\linewidth}
      \centering\includegraphics[width=1\linewidth]{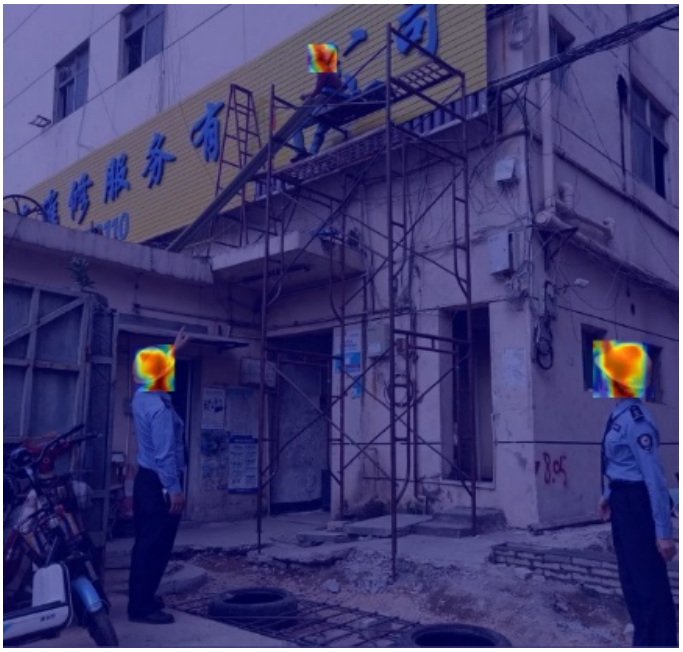}\vspace{0.04cm}
      \centering\includegraphics[width=1\linewidth]{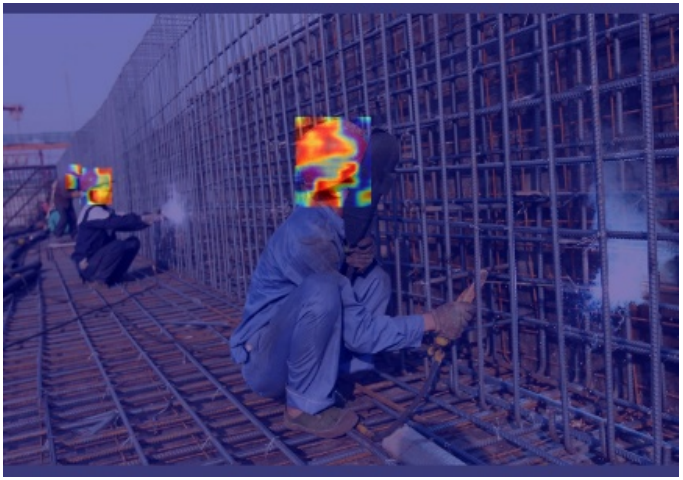}\vspace{0.04cm}
      \centering\includegraphics[width=1\linewidth]{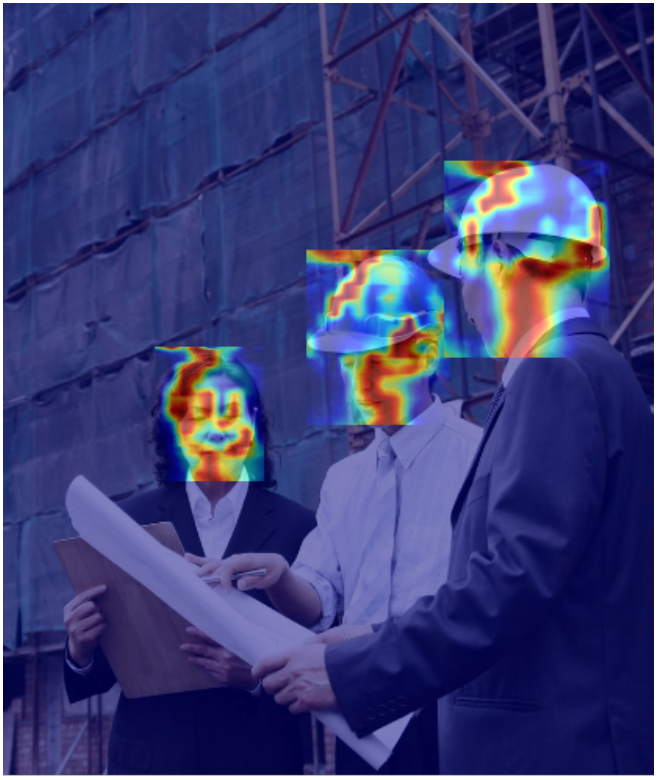}
    \end{minipage}         
}         
\caption{Heatmap comparison of detection results based on Grad-CAM\cite{60}. YOLOv3-tiny, YOLOv8, and DABFNet attention maps on different scenes, showing model focus areas.}
\label{fig6}
\end{figure}

\begin{table*}[t]
\centering
\caption{Comparison of the improved algorithm with five baseline algorithms on different categories of data on the safety helmet wearing dataset.}
\label{tab:3} 
\begin{tabular}{cccccccccccc}
\hline
\multirow{2}{*}{\textbf{Model}} & \multicolumn{5}{c}{Average}                                                  & \textbf{} & \multicolumn{2}{c}{Hat}       & \textbf{} & \multicolumn{2}{c}{Person}    \\ \cline{2-6} \cline{8-9} \cline{11-12}
                                & Precision             & Recall             & mAP@0.5       & mAP@[0.5:0.95]      & GFLOPs       & \textbf{ } & Recall             & mAP@0.5        & \textbf{ } & Recall             & mAP@0.5        \\ \hline
YOLOv3-tiny\cite{49}                     & 90.0          & 77.1          & 85.0          & 54.4          & 19.0         &           & 87.8          & 91.4          &           & 66.5          & 78.5          \\
YOLOv5\cite{1}                          & 92.4          & 88.5          & 93.6          & 60.4          & \textbf{7.1} &           & 88.8          & 93.8          &           & 88.1          & 93.3          \\
YOLOv5-p6\cite{50}                       & 93.5          & 88.1          & 93.5          & 60.6          & 7.2          &           & 88.9 & 93.6          &           & 87.4          & 93.3          \\
YOLOv6\cite{51}                          & 92.4          & 86.9          & 93.0          & 60.3          & 11.8         &           & 88.0          & 93.5          &           & 85.9          & 92.4          \\
YOLOv8\cite{38}                          & 93.1          & 87.7          & 93.7          & 60.9          & 8.1          &           & 88.4          & 94.2          &           & 87.1          & 93.1          \\
DABFNet (Ours)                & \textbf{93.5} & \textbf{89.8} & \textbf{94.9} & \textbf{62.6} & 9.0          &           & \textbf{90.2}          & \textbf{95.3} &           & \textbf{89.3} & \textbf{94.4} \\ \hline
\end{tabular}
\vspace{-3mm}
\end{table*}

 \begin{table}[t]
 \centering
 \caption{Ablation experiment results of the improved algorithm. We use mAP@{[}0.5:0.95{]} as the evaluation metric.}
 \label{tab:9} 
 \setlength{\tabcolsep}{0.3mm}{
 \begin{tabular}{cccccccccc}
 \hline
 DAHead & BWFPN & WIoU  & Precision & Recall & mAP@0.5 & mAP@{[}0.5:0.95{]} & GFLOPs  \\ \hline
 \ding{55}        &\ding{55}       &\ding{55}                            & 93.1      & 87.7   & 93.7    & 60.9               & 8.1       \\
\ding{51}        &\ding{55}       &\ding{55}                    & 93.4      & 88.3   & 94.1    & 61.3               & 9.6       \\
                                  \ding{55}        &\ding{51}       &\ding{55}                     & 92.7      & 88.7   & 93.8    & 61.1               & \textbf{7.1}       \\
                                  \ding{55}        &\ding{55}       &\ding{51}                     & 92.9      & 89.4   & 93.8    & 60.8               & 8.1       \\
                                  \ding{51}        &\ding{51}       &\ding{55}              & 93.0      & 88.0   & 93.8    & 61.3               & 9.0       \\
                                  \ding{51}        &\ding{55}       &\ding{51}              & \textbf{93.7}    & 89.1   & 94.1    & 61.4               & 9.6       \\
                                  \ding{55}        &\ding{51}       &\ding{51}              & 92.4      & 89.3   & 93.8    & 61.1               & 7.1      \\
                                  \ding{51}        &\ding{51}       &\ding{51}           & 93.5      & \textbf{89.8}   & \textbf{94.9}    & \textbf{62.6}               & 9.0      \\ \hline
 \end{tabular}}
 \vspace{-3mm}
 \end{table}

 \subsection{\textbf{Comparison Experiment of Different Sizes}}
 The main difference between the n, s, m, l, x versions of the model size is the size and complexity of the model, which affects the performance of the model and the use of computing resources. As shown in Fig.~\ref{fig7}, the accuracy, recall, mAP@0.5 and mAP@[0.5:0.95] of the YOLOv8 and DABFNet models are improved as the models become deeper. In general, the accuracy, recall, mAP@0.5 and mAP@[0.5:0.95] of the DABFNet proposed by us are better than those of YOLOv8 at the same size. In addition, DABFNet has lower GFLOPs under the same size,which has an advantage in the hardware deployment of the model.


 \subsection{\textbf{Visualization}}
 To demonstrate the accuracy of the improved algorithm proposed in this paper in different scenarios, we show the detection effects of YOLOv3-tiny, YOLOv8 and our DABFNet in three scenarios. Fig.~\ref{fig5} directly shows the detection effect of the model, intuitively comparing the performance of each model in different scenarios; Fig.~\ref{fig6} uses heat maps to show the degree of attention of the three models on the target area. From the results, it can be seen that in the first scene, both YOLOv3-tiny and YOLOv8 have some misdetection of the background, while the improved model accurately detects all targets. The buildings in the construction scene are complex and their colors are similar to those of the safety helmets. It is very easy for the model to mistakenly identify the background as workers wearing safety helmets. Therefore, the first scene verifies the good robustness of our model. In the second scene, there is a row of workers working. YOLOv3-tiny still misdetects the background as someone wearing a helmet. Although YOLOv8 does not misdetect the background, it misses the worker farthest away. The improved algorithm in this paper achieves a balance between the two. Not only does it avoid false detections, but it also accurately detects small targets in the distance. Therefore, the second scenario verifies the effectiveness of the proposed algorithm for detecting small targets and occluded targets. In the third scenario, there are four workers in one picture. YOLOv3-tiny obviously mistakenly detects the incomplete portrait on the far right as wearing a helmet. Compared with the detection result of YOLOv8, the detection frame of the improved YOLOv8 algorithm is more reasonable and has a higher confidence level. This verifies the accuracy of the improved algorithm.

\section{CONCLUSION}
To address the challenges of detecting safety helmet usage in complex environments, such as construction sites with frequent occlusions and small target detection issues, this paper enhances the YOLOv8 baseline model and proposes an improved algorithm for safety helmet detection. An attention module is integrated into the convolutional layers to leverage the attention mechanism for aggregating critical information from feature maps, thereby enhancing the feature extraction capabilities of the CNN. The traditional FPN feature fusion structure is replaced with Balanced Weighted Feature Pyramid Network, which improves the efficiency of fusing feature maps across different scales and resolutions. Additionally, the CIoU loss function is replaced with Wise-IoUv3 to enable finer and more effective optimization, further boosting the model's detection performance.
\textbf{Future Work.} The proposed algorithm achieves a 94.9\% mAP on the test set while maintaining real-time detection speed, establishing state-of-the-art results in safety helmet detection. Despite these advancements, the algorithm faces challenges with false detections. Future work will focus on reducing the model's sensitivity to background features to minimize false positives, making the approach more robust and applicable to real-world scenarios.

\bibliographystyle{IEEEtran}
\bibliography{reference}

\end{document}